\documentclass[11pt,a4paper]{article}
\usepackage[hyperref]{emnlp2020}
\usepackage{times}
\usepackage{latexsym}

\usepackage{url}

\usepackage{graphicx}
\usepackage{subcaption}
\usepackage{amsmath,bm}
\usepackage{mathrsfs}
\usepackage{enumitem}
\usepackage{framed}
\usepackage{array}
\usepackage{multirow}
\usepackage{enumitem}
\usepackage{amsfonts}

\newcommand{\PreserveBackslash}[1]{\let\temp=\\#1\let\\=\temp}
\newcolumntype{C}[1]{>{\PreserveBackslash\centering}p{#1}}
\newcolumntype{R}[1]{>{\PreserveBackslash\raggedleft}p{#1}}
\newcolumntype{L}[1]{>{\PreserveBackslash\raggedright}p{#1}}

\newtheorem{conclusion}{Conclusion}
\usepackage{colortbl}
\usepackage{lipsum}
\definecolor{mygray}{gray}{.9}
\definecolor{mypink}{rgb}{.99,.91,.95}
\definecolor{mycyan}{cmyk}{.3,0,0,0}
\definecolor{green}{RGB}{0,100,0}
\usepackage{stmaryrd}
\usepackage{mathabx}

\aclfinalcopy 

\title{A Rigorous Study on Named Entity Recognition: Can Fine-tuning Pretrained Model Lead to the Promised Land?}

\author{Hongyu Lin${}^{1}$, Yaojie Lu${}^{1,3}$, Jialong Tang${}^{1,3}$, Xianpei Han${}^{1,2,*}$ , Le Sun${}^{1,2,*}$ \\ {\bf Zhicheng Wei}${}^{4}$ , {\bf Nicholas Jing Yuan}${}^{4}$ \\
${}^{1}$Chinese Information Processing Laboratory ~ ${}^{2}$State Key Laboratory of Computer Science \\
Institute of Software, Chinese Academy of Sciences, Beijing, China\\
${}^{3}$University of Chinese Academy of Sciences, Beijing, China \\
${}^{4}$Huawei Cloud\&AI\\
 {\{hongyu,yaojie2017,jialong2019,xianpei,sunle\}@iscas.ac.cn} \\
 {\{weizhicheng1,nicholas.yuan\}@huawei.com}
}

\date{}

\begin{document}
\maketitle

\newcommand\blfootnote[1]{%
\begingroup
\renewcommand\thefootnote{}\footnote{#1}%
\addtocounter{footnote}{-1}%
\endgroup
}

\begin{abstract}
  Fine-tuning pretrained model has achieved promising performance on standard NER benchmarks. 
  Generally, these benchmarks are blessed with strong name regularity, high mention coverage and sufficient context diversity. Unfortunately, when scaling NER to open situations, these advantages may no longer exist. And therefore it raises a critical question of whether previous creditable approaches can still work well when facing these challenges. As there is no currently available dataset to investigate this problem, this paper proposes to conduct randomization test on standard benchmarks. Specifically, we erase name regularity, mention coverage and context diversity respectively from the benchmarks, in order to explore their impact on the generalization ability of models. To further verify our conclusions, we also construct a new open NER dataset that focuses on entity types with weaker name regularity and lower mention coverage to verify our conclusion. From both randomization test and empirical experiments, we draw the conclusions that 1) name regularity is critical for the models to generalize to unseen mentions; 2) high mention coverage may undermine the model generalization ability and 3) context patterns may not require enormous data to capture when using pretrained encoders.\blfootnote{*: Corresponding authors.}
\end{abstract}

\section{Introduction}
Named entity recognition (NER), or more generally name tagging, aims to identify text spans pertaining to specific entity types. NER is a fundamental task of information extraction which enables many downstream NLP applications, such as relation extraction~\cite{guodong2005exploring,mintz2009distant}, event extraction~\cite{ji2008refining,li2013joint} and machine reading comprehension~\cite{rajpurkar2016squad,wang2016multi}. 
Recently, neural network-based supervised models dominate the NER task. By supervised fine-tuning upon large-scale language model pretrained architectures (e.g., ELMo~\cite{peters2018deep}, BERT~\cite{devlin2018bert}, XLNet~\cite{yang2019xlnet}, etc.), we have witnessed superior performances on almost all widely-used NER benchmarks, including CoNLL03, ACE2005 and TAC-KBP datasets~\cite{li2019entity,akbik2019pooled,zhai2019improving,li2019unified}. 

\begin{figure}[!t]
  \centering
  \includegraphics[width=0.49\textwidth]{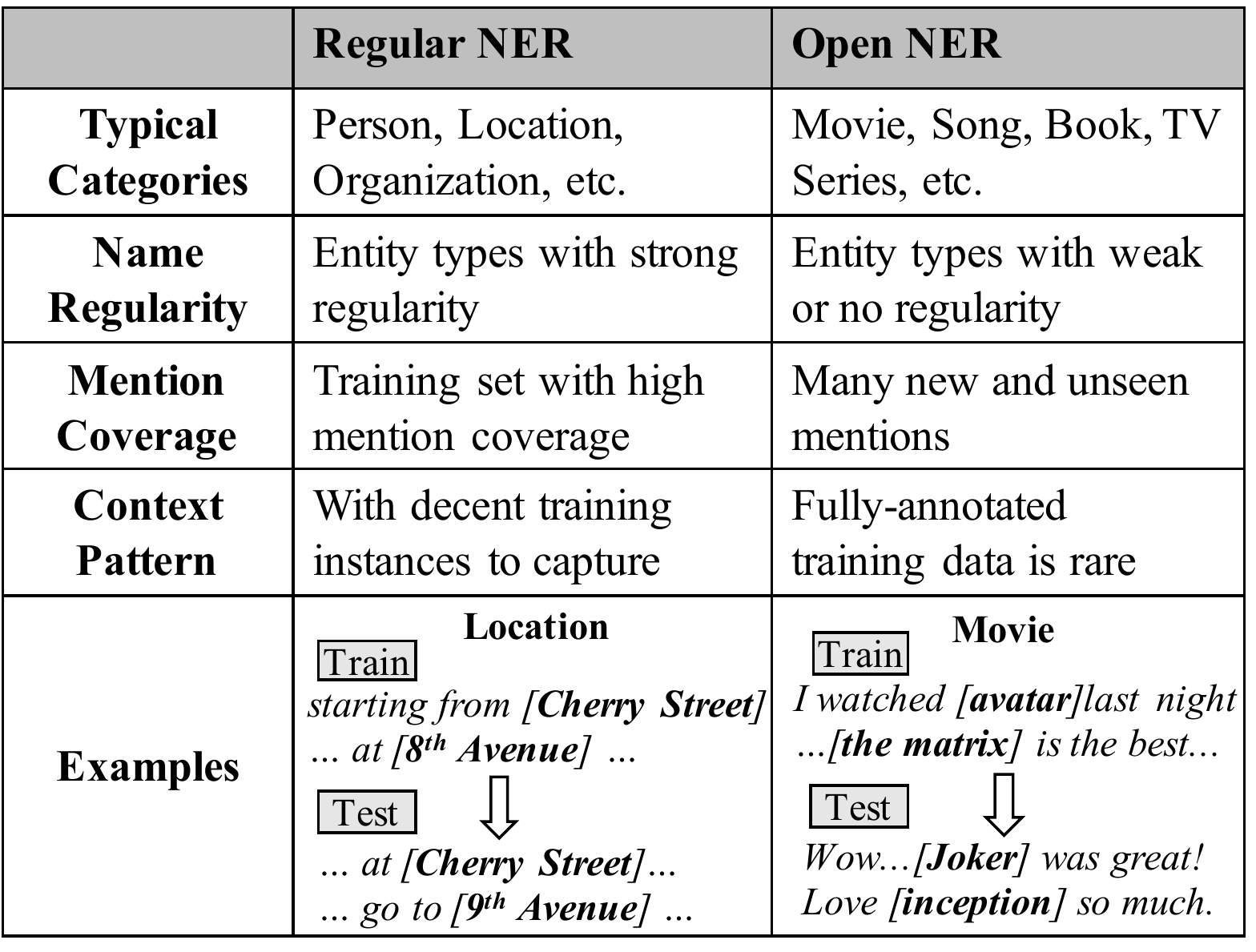}\\
  \caption{Comparison between regular NER benchmarks and open NER tasks in reality.}\label{tab:reg_open_ner}
\end{figure}

Despite the success of recent models, there are specific advantages in current NER benchmarks which significantly facilitate supervised neural networks. First, these benchmarks focus on limited entity types, and most mentions of these types have strong \emph{\textbf {name regularity}}.  For example, nearly all person names follow the ``FirstName LastName'' or ``LastName FirstName'' patterns, while location names mostly end with indicator words such as ``street'' or ``road''.  
Second, the training and test data in these benchmarks are sampled from the same corpus, and therefore the training data usually have high \emph{\textbf {mention coverage}} on the test data, i.e., a large proportion of mentions in the test set have been observed in the training set. However, it is obvious that this high coverage is inconsistent with the primary goal of NER models, which is expected to identify unseen mentions from new data by capturing the generalization knowledge about names and contexts. For observed mentions, other techniques, such as entity linking~\cite{lin2012entity}, would be more appropriate and effective. Third, these benchmarks generally provide decent training data, and therefore the \emph{\textbf {context diversity}} of all entity types can be sufficiently learned. In this paper, we refer to the NER tasks with strong name regularity, high mention coverage and with sufficient training instances as \emph{regular NER}. And it proves that the state-of-the-art neural networks can easily exploit such name regularity, mention coverage and context diversity knowledge, and therefore achieve state-of-the-art performance in these benchmarks.

Unfortunately, when it comes to a more general scenario, there are significant discrepancies between regular benchmarks and NER in open settings. Table~\ref{tab:reg_open_ner} overviews their discrepancies on name regularity, mention coverage and context pattern acquisition. Specifically, mentions of many entity types do not follow regular compositional structures. For example, a movie name can be any n-gram utterance and even is not a regular noun phrase(e.g., ``Gone with the Wind''). Furthermore, fully-annotated training data will be rare due to the expensive cost. As a consequence, training set can only cover a minor part of test mentions and diverse context patterns must be learned from minimal instances. It is obvious that these discrepancies will lead to the biased estimation of the open NER performance using regular NER benchmarks.

In this paper, we want to shed some light on the impact of the discrepancies between regular and open NER, and provides some valuable insights into the construction of general NER models in a more effective and efficient way. Specifically, we want to answer the following question:

\emph{\textbf{Can pretrained supervised neural networks still generalize well on NER when either weaker name regularity, lower mention coverage or inadequate context diversity exists?}}

It is non-trivial to answer this question because currently no well-established benchmark concentrates on these issues. To this end, this paper exploits the efficacy of the above three kinds of information by conducting a series of experiments based on \emph{randomization test}~\cite{edgington2007randomization,zhang2016understanding}. Specifically, we design several mention replacing mechanisms, which can erase specific kinds of information on-demand from current NER benchmarks. By applying the same supervised models on both vanilla and information-erased data, we can investigate how much the models will rely on particular erased information to identify entity mentions. Generally, we propose to erase name regularity, mention coverage and context diversity respectively using the following kinds of randomization test, whose examples are shown in Table~\ref{tab:rep_instances}:

\begin{table*}[!ht]
  \small
  \centering
  \begin{tabular}{|L{2.8cm}|C{1.0cm}|C{1.0cm}|C{1.0cm}|L{7.3cm}|}
    \hline
    \rowcolor{mygray} {\bf Settings} & {\bf Name} & {\bf Mention} & {\bf Context} & {\bf Examples}\\
    \hline
    Vanilla Baseline & $\surd$ & $\surd$ & $\surd$
    & 
      $
      \mbox{Train} \left\{
      \begin{aligned}
      &\mbox{ \emph{[\textcolor{red}{Putin}] concluded his two days of talks.}}\\
      &\mbox{ \emph{[\textcolor{blue}{Blair}] spoke to [\textcolor{green}{Bush}] on April 5. }}
      \end{aligned}
      \right.
      $
      \par \ \par
      \ Test\quad \   \emph{[\textcolor{red}{Putin}] will face re-election in March 2004. }
    \\ \hline
    
    Name Permutation (NP) & $\bigtimes$ & $\surd$ & $\surd$
    & 
    $
    \mbox{Train} \left\{
    \begin{aligned}
    &\mbox{ \emph{[\textcolor{red}{the united}] concluded his two days of talks.}}\\
    &\mbox{ \emph{\textcolor{blue}{[Hillsborough}] spoke to [\textcolor{green}{analysts}] on April 5. }}
    \end{aligned}
    \right.
    $
    \par\ \par
    \ Test\quad \   \emph{[\textcolor{red}{the united}] will face re-election in March 2004. }
    \\ \hline
    
    Mention Permutation (MP) & $\bigtimes$ & $\bigtimes$ & $\surd$
    &
    $
    \mbox{Train} \left\{
    \begin{aligned}
    &\mbox{ \emph{[\textcolor{red}{the united}] concluded his two days of talks.}}\\
    &\mbox{ \emph{\textcolor{blue}{[Hillsborough}] spoke to [\textcolor{green}{analysts}] on April 5. }}
    \end{aligned}
    \right.
    $
    \par \ \par
    \ Test\quad \   \emph{[\textcolor{red}{which girl}] will face re-election in March 2004. }
    \\ \hline

    \par Context Reduction (CR) & $\surd$ & $\surd$ & $\downarrow$
    &
    $
    \mbox{Train} \left\{
    \begin{aligned}
    &\mbox{ \emph{\textcolor{red}{[Putin}] concluded his two days of talks.}}\\
    &\mbox{ \emph{[\textcolor{blue}{Blair}] concluded his two days of talks. }} \\
    &\mbox{ \emph{[\textcolor{green}{Bush}] concluded his two days of talks. }} \\
    \end{aligned}
    \right.
    $
    \par \ \par
    \ Test\quad \   \emph{[\textcolor{red}{Putin}] will face re-election in March 2004. }
    \\ \hline

    \par Mention Reduction (MR) & $\downarrow$ & $\downarrow$ & $\surd$
    &      
    $
    \mbox{Train} \left\{
    \begin{aligned}
    &\mbox{ \emph{[\textcolor{blue}{Blair]} concluded his two days of talks.}}\\
    &\mbox{ \emph{[\textcolor{blue}{Blair}] spoke to [\textcolor{blue}{Blair}] on April 5. }}
    \end{aligned}
    \right.
    $
    \par \ \par
    \ Test\quad \   \emph{[\textcolor{red}{Putin}] will face re-election in March 2004. }
    \\ \hline
    
  \end{tabular}
  \caption{\label{tab:rep_instances}Illustration of our four kinds of randomization test. The utterances in square brackets are entity mentions. Name: name regularity knowledge; Mention: high mention coverage; Context: sufficient training instances for context diversity $\surd$: the knowledge is preserved in this setting; $\bigtimes$: the knowledge is erased from the data in the setting; $\downarrow$: the knowledge decreases.}
  \end{table*}

\begin{itemize}[itemsep=0pt,leftmargin=10pt,topsep=3pt]
  \item \textbf{Name Permutation (NP)} is used to investigate the necessity of name regularity for NER, which replaces the same entity mention with an identical, random n-gram string. In this way, the structural correlation between mentions of the same type is removed. For the example in Table~\ref{tab:rep_instances}, all mention ``Putin'' is replaced by the same utterance ``the united''.

  \item \textbf{Mention Permutation (MP)} is used to investigate the impact of mention coverage. Different from NP, MP replaces each mention with a unique n-gram string, and even two mentions with the same utterance will be replaced by different strings. For the example in Table~\ref{tab:rep_instances}, two mentions of ``Putin'' are replaced by ``the united'' and ``which girl'' respectively. In this way, the mention coverage is erased and the model should merely rely on context knowledge for NER prediction.

  \item \textbf{Context Reduction (CR)} and \textbf{Mention Reduction (MR)} are used to investigate the influence of less training data. CR decreases the diversity of sentences but preserves all mentions in vanilla data, while MR keeps all sentences but only preserves a small part of the original mentions. We compare these two settings, to figure out how much original training data are needed to learn context patterns and name regularity.
\end{itemize}

To verify the above findings, we further conduct a verification experiment by constructing a new dataset derived from Wikipedia, which focuses on entity types with weak name regularity. To the best of our knowledge, this is the first work that tries to investigate such critical differences between regular and open NER. From both the randomization test and the verification experiment, we reach the following main conclusions:

\begin{itemize}[itemsep=0pt,leftmargin=10pt,topsep=0pt]
  \item \emph{\textbf{Decent name regularity is vital to the generalization over unseen entity mentions.}} When name regularity is erased, the performance on unseen mentions will be significantly undermined. This finding indicates that it will be challenging to build models for open entity types with weak name regularity.

  \item \emph{\textbf{High mention coverage weakens the model ability to capture informative context knowledge.}} In other words, high mention coverage will mislead models to memorize popular mentions, rather than to learn generalization knowledge. This also reveals that current performance on regular NER benchmarks is highly biased, i.e., the performance on open NER will be significantly lower than that on regular benchmarks.

  \item \emph{\textbf{Sufficient context diversity may not require enormous training data to capture.}} We show that with simple data augmentation techniques to preserving name regularity, required training data can be significantly reduced. This observation also raises the possibility of designing more effective NER models with less annotated data.
\end{itemize}

\section{Experiment Settings}

\subsection{Dataset Summary}
We use ACE2005 (LDC2006T06) as our primary experiment dataset for randomization test. Other openly-available datasets, such as CoNLL03 and Ontonotes, are not suitable for our randomization test. This is because they only annotate named mentions but ignore nominal and pronominal mentions. However, the context of named and nominal/pronominal mentions is generally identical, and therefore the models will be unable to distinguish between them once name regularity is removed.

For better illustration and reproduction, we will report experiment results using the same dataset splits corresponding to ~\citet{D18-1124,D18-1019,lin2019sequence,xia2019multi}. For all experiments, we only consider the outmost mentions similar to the majority of the previous work. Finally, there are 18739/2531/2314 mentions in the train/dev/test set respectively. We found that 58.4\% mentions in the test set have appeared in the training data, which confirms our high mention coverage concern. We also have conducted multiple experiments using the 8:1:1 train/dev/test data split. And we found all the above experiments lead to the same conclusions which we will illustrate in the next section.

\begin{table*}[!ht]
  \centering
  \resizebox{0.95\textwidth}{!}{
    \begin{tabular}{l|ccccccc|c}
    \hline
    \textbf{Data Setting} & \textbf{PER} & \textbf{ORG} & \textbf{GPE}  & \textbf{FAC} 
    & \textbf{LOC}  & \textbf{WEA} & \textbf{VEH} & \textbf{ALL} \\
    \hline\hline
    Baseline                          & 86.31   & 76.49    & 80.89   & 69.23   & 40.58   & 74.70 & 61.97  & 81.76   \\
    \hline

    Name Permutation                  & 73.41    & 44.34   & 49.71  & 37.96    & 28.24   & 33.33 & 23.93 & 62.28   \\
    - Drop Compared with Baseline     & 15\%     & 42\%    & 39\%   & 45\%     & 44\%   & 55\% & 61\% & 24\%   \\
    \hline
    Mention Permutation & 61.78    & 39.40   & 33.27  & 32.16    & 18.60   & 9.38   & 21.92 & 51.58   \\
    - Drop Compared with Baseline     & 28\%    & 48\%   & 59\%     & 54\%    & 54\%   & 87\% & 65\% & 34\%   \\
    \hline
    \end{tabular}
  }
    \caption{Micro-F1 scores of BERT-CRF tagger on original data, name permutation setting and mention permutation setting respectively. We can see that erasing name regularity and mention coverage will significantly undermine the model performance.}
  \label{tab:overall_rst}%
\end{table*}

\subsection{Baseline}
We use the BERT-CRF tagger as our baseline. Specifically, a Transformer~\cite{vaswani2017attention} is used as the encoder and then two dense layers are used to map the hidden representation into the label space. Finally, a linear-chain CRF is applied. The transformer is initialized using \emph{bert\_uncased\_L-24\_H-1024\_A-16}, which achieves the best performance on our auxiliary experiments.  All model parameters are fine-tuned later. We used Adam~\cite{kingma2014adam} as the optimizer and set the learning rate to $10^{-5}$. Finally, this model achieves 81.76 micro F1 score on ACE2005.

\section{Randomization Test on NER}
To investigate the discrepancies between regular and open NER, this paper controllably erases target information from the vanilla data via a variant of randomization test~\cite{edgington2007randomization,zhang2016understanding} in non-parametric statistics. Concretely, to probe the effect of a specific kind of information, we erase it in vanilla data by randomly replacing entity mentions with particular irregular utterances. After that, we learn and compare NER models on both the vanilla benchmark and the information-erased version, in order to evaluate the models’ robustness and generalization ability when target information is absent. The results of our randomization test can serve as a frame of reference for open NER, where the erased information is often truly absent.

Specifically, three kinds of information are particularly considered, and four kinds of strategies are used in our randomization test. Table~\ref{tab:rep_instances} shows all our randomization test with examples. In the following, we will illustrate the empirical findings through the test, with one subsection for one kind of information. For each kind of information, we first present the critical conclusion and then demonstrate how we reach the conclusion.

\subsection{Effect of Name regularity}
\label{sec:name_regularity}
\begin{conclusion}
Name regularity is critical for supervised NER model to generalize over unseen entity mentions.
\end{conclusion}

One critical difference between regular and open NER is whether names of the same entity type share inner compositional structure. In regular NER, entities (e.g., PER, ORG and LOC) have long been observed with strong name regularity. In open NER, however, most entities (e.g., movie, song and book) do not have such strong name regularity, and some of mentions can even be random utterances. Therefore, it is critical to evaluate the impact of name regularity on generalization.

To address this issue, we propose \emph{name permutation}, which replaces each mention utterance with a randomly sampled n-gram string, and the mentions with the same name will all be replaced by the same string. To ensure that no structural correlation between these mentions will be retained, the replacing strings are randomly sampled. For example in Table~\ref{tab:rep_instances}, all mentions ``Putin'' are replaced by ``the united'', and all ``Bush'' are replaced by ``analysts``'. In this way, the name regularity will be erased, but the mention coverage will still retain because the same mention in the training and test data will still be the same.

Table~\ref{tab:overall_rst} shows the overall results. We can see that when we erase name regularity from the dataset, the performance significantly undermined. The overall drop on micro-F1 is 24\%. Moreover, in the majority of entity types, the performance slips more than 40\%. This verifies the importance of name regularity on model generalization ability. To investigate the reasons behind, we split mentions for evaluation by whether the predicted/golden mention is covered by the training data, which we refer to as the in-dictionary portion (InDict) and the out-of-dictionary portion (OutDict) respectively. The results~\footnote{We only present the performance on four entity types with sufficient training and testing instances.} of these two portions on the vanilla dataset and name permutation setting are shown in Table ~\ref{tab:full_np}.

\begin{table}[!ht]
  \centering
  \resizebox{0.48\textwidth}{!}{
    \begin{tabular}{l|cc|c||cc|c}
    \hline
    \multicolumn{7}{c}{\textbf{Vanilla Baseline}} \\
    \hline
    & \multicolumn{3}{c||}{Precison} & \multicolumn{3}{c}{Recall} \\
    \cline{2-7}
    & InDict & OutDict & \textbf{Diff} &  InDict & OutDict & \textbf{Diff} \\
    \hline
    PER	&	88.03	&	75.40	&	14\%	&	92.90	&	85.20	&	8\% \\
    ORG	&	73.51	&	72.77	&	1\%	  &	81.93	&	76.56	&	7\% \\
    GPE	&	79.55	&	78.21	&	2\%	  &	85.37	&	77.22	&	10\% \\
    FAC	&	65.91	&	65.67	&	0\%	  &	86.05	&	65.67	&	24\% \\
    \hline
    ALL	&	83.37	&	72.97	&	\textbf{12\%}	&	89.08	&	79.11	&	\textbf{11\%} \\
    \hline 
    \hline
    \multicolumn{7}{c}{\textbf{Name Permutation}} \\
    \hline
    & \multicolumn{3}{c||}{Precison} & \multicolumn{3}{c}{Recall} \\
    \cline{2-7}
    & InDict & OutDict & \textbf{Diff} &  InDict & OutDict & \textbf{Diff} \\
    \hline
    PER	&	88.58	&	46.91	&	47\%	&	87.00	&	62.30	&	28\% \\
    ORG	&	70.40	&	37.01	&	47\%	&	51.76	&	27.80	&	46\% \\
    GPE	&	70.20	&	18.60	&	74\%	&	64.63	&	30.38	&	53\% \\
    FAC	&	63.64	&	27.40	&	57\%	&	48.84	&	29.85	&	39\% \\
    \hline
    ALL	&	82.47	&	38.29	&	\textbf{54\%}	&	76.31	&	46.72	&	\textbf{39\%} \\
    \hline
    \end{tabular}
  }
    \caption{
      Comparasion between baseline and name permutation on in-dictionary and out-of-dictionary portions. We can see that the performance gap between InDict and OutDict is significantly enlarged when name regularity was erased.}
  \label{tab:full_np}%
\end{table}

From Table~\ref{tab:full_np}, we can find that erasing name regularity leads to more severe performance drop on mentions not covered by training set (OutDict) than those appearing in the training set (InDict). For the vanilla dataset, the performance gap between InDict and OutDict is not very large, which shows the good generalizing ability of pretrained supervised model over unseen mentions with strong name regularity. However, after erasing name regularity, this gap is significantly enlarged. The performance on the InDict portion does not drop too much, but the performance on the OutDict portion drops dramatically. This result shows that it is quite difficult to recognize unseen entity mentions when name regularity is missing. Besides, we can see that after erasing name regularity, the model can still perform quite well on the in-dictionary portion, whose precision is still quite high. This demonstrates the strong ability of neural networks to memorize and disambiguate observed mentions even their names are irregular.

Note that Name permutation may result in ambiguous entity instances with only sentence contexts. However, this phenomenon is very common in real world, open-ended NER tasks. For example, a twitter is often a simple sentence like “La La Land is great, we like it”, where “La La Land” is ambiguous in this context and can only be recognized as a Movie using world knowledge about movie. Our experiments also confirm that the less context provides, the more name knowledge is needed.

To summarize, name regularity is very critical for model generalization over unseen mentions. Without name regularity, current models can only work well on mentions covered by training data via memorizing and disambiguating names, but cannot generalize well to unseen mentions.

\subsection{Effect of Mention Coverage}
\begin{conclusion}
  High mention coverage weakens the models' ability of capturing informative generalization knowledge for NER.
  \end{conclusion}

  Another critical difference between regular and open NER is whether the training data can cover a majority of mentions in the test scenario. High mention coverage can provide misleading evidence during model learning because neural networks can achieve considerable performance by just memorizing and disambiguating observed entity names. This ability, obviously, is not what we desire because 1) in real world applications, most entity mentions are new and unseen, which means out-of-dictionary mentions will dominate the test process; 2) because the training instances are very limited in open situations, it is too expensive to achieve high mention coverage; 3) many long-tail mentions in the training set would be one-shot, i.e., the mention only appears once in the training data. Therefore, it is necessary to exploit whether NER models can still reach reasonable performance in low mention coverage situation.

  To this end, we conducted experiments via \emph{mention permutation}, which replaces each mention with a random n-gram similar to the name permutation. However, to erase mention coverage information, the replacing string for each mention is independently sampled, and therefore even mentions with the same utterance in vanilla data will be replaced by different strings. For example, two ``Putin'' in Table~\ref{tab:rep_instances} are replaced by different utterances. In this way, (almost) no mention in the test set is covered by the training data, and no name information remains in the data. Consequently, the models should only rely on context knowledge for NER prediction.
  
  The mention permutation results are shown in Table~\ref{tab:overall_rst}. We can see that the performance of MP further drops compared with NP, which demonstrates high mention coverage can make mention detection much simpler. To further investigate whether high mention coverage will influence the models’ generalization ability, we also compared MP with NP on the \emph{out-of-dictionary portion}. The results are shown in Table~\ref{tab:rp_np}. 
  
  \begin{table}[!t]
    \centering
    \resizebox{0.35\textwidth}{!}{
      \begin{tabular}{l|cc|cc}
      \hline
      & \multicolumn{2}{c|}{Precision} & \multicolumn{2}{c}{Recall}  \\
      \cline{2-5}
       & NP & MP & NP &  MP\\
      \hline
      PER	&	46.91	&	\textbf{58.01}	&	62.30	&	\textbf{66.21} \\
      ORG	&	37.01	&	\textbf{40.76}	&	27.80	&	\textbf{38.94} \\
      GPE	&	18.60	&	\textbf{32.92}	&	30.38	&	\textbf{34.05} \\
      FAC	&	27.40	&	\textbf{27.74}	&	29.85	&	\textbf{36.54} \\
      \hline
      ALL	&	38.29	&	\textbf{49.40}	  &	46.72	&	\textbf{54.01} \\
      
      \hline
      
      \end{tabular}
    }
      \caption{
      Experiment results on \emph{OutDict} portion. We can see that mention permutation significantly performs better than name permutation, which indicates that high mention coverage may undermine the generalization ability of models.}
    \label{tab:rp_np}%
  \end{table}

  Surprisingly, the model performs significantly better in the MP setting than in the NP setting in all entity types. In other words, high mention coverage undermines the models’ ability to generalize to unseen mentions. We believe this is because, as some previous studies in other tasks~\cite{zhang2016understanding,lu2019distilling} have pointed out, neural networks have strong ability and tendency to memorize training instances. Consequently, the high mention coverage will mislead the models to mainly memorize and disambiguate frequent entity names even though they are irregular, but ignore informative context patterns which are useful for generalization over unseen mentions. These results reveal that NER models should focus more on context knowledge for generalization, rather than memorizing popular mentions. This is even more important for entity types without or with weak name regularity because context patterns are more critical in this circumstance.

\subsection{Effect of Context Diversity}

\begin{conclusion}
  \label{con:3}
  Sufficient context patterns may not require enormous training data to capture when learning upon pretrained neural networks.
  \end{conclusion}

  \begin{figure*}[!ht]
    \centering
    \resizebox{1.00\textwidth}{!}{
      \begin{subfigure}[c]{0.33\textwidth} 
        \centering
        \includegraphics[width=\textwidth]{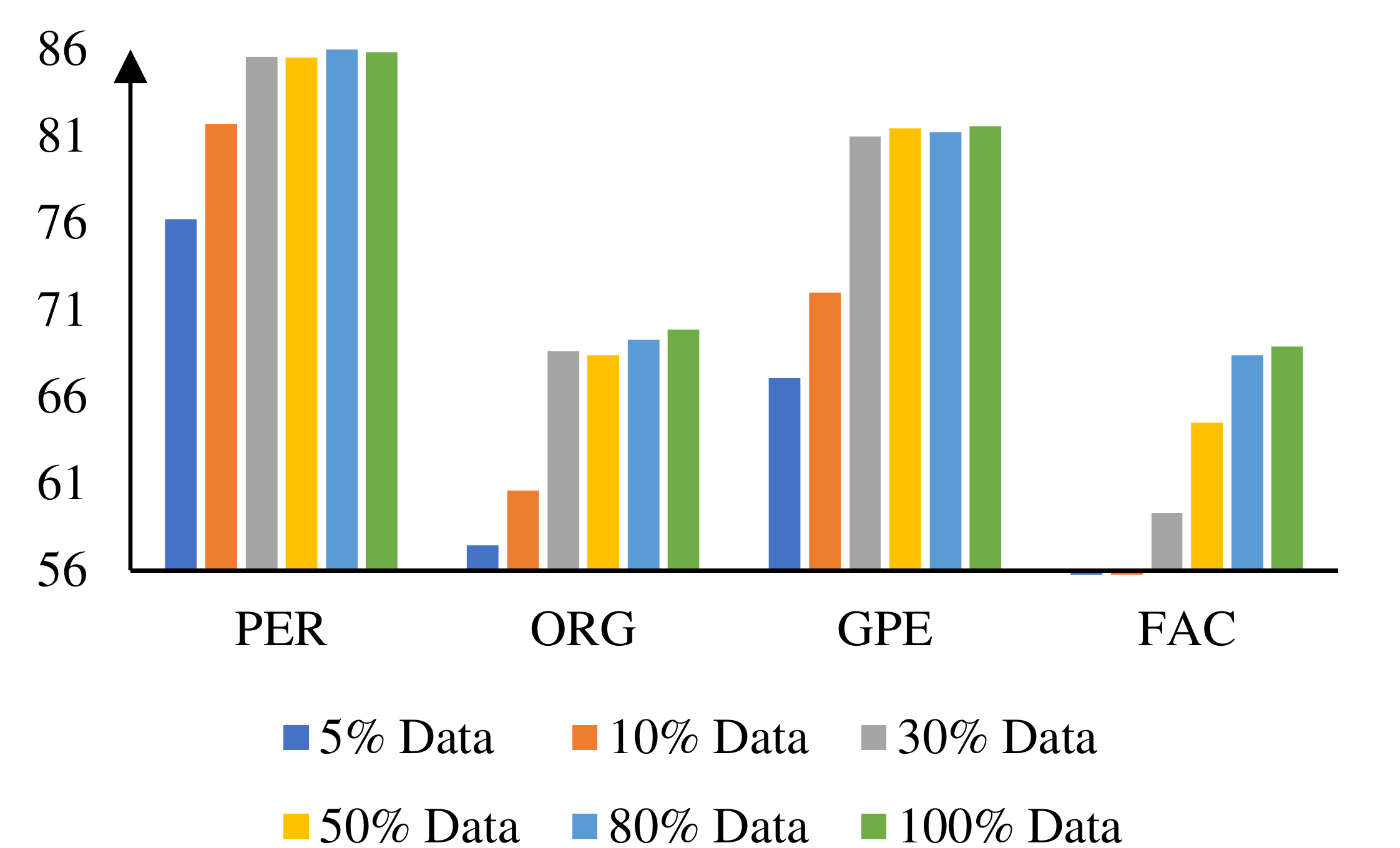} 
        \caption{Context Reduction}
      \end{subfigure}%
      \begin{subfigure}[c]{0.33\textwidth}
        \centering
        \includegraphics[width=\textwidth]{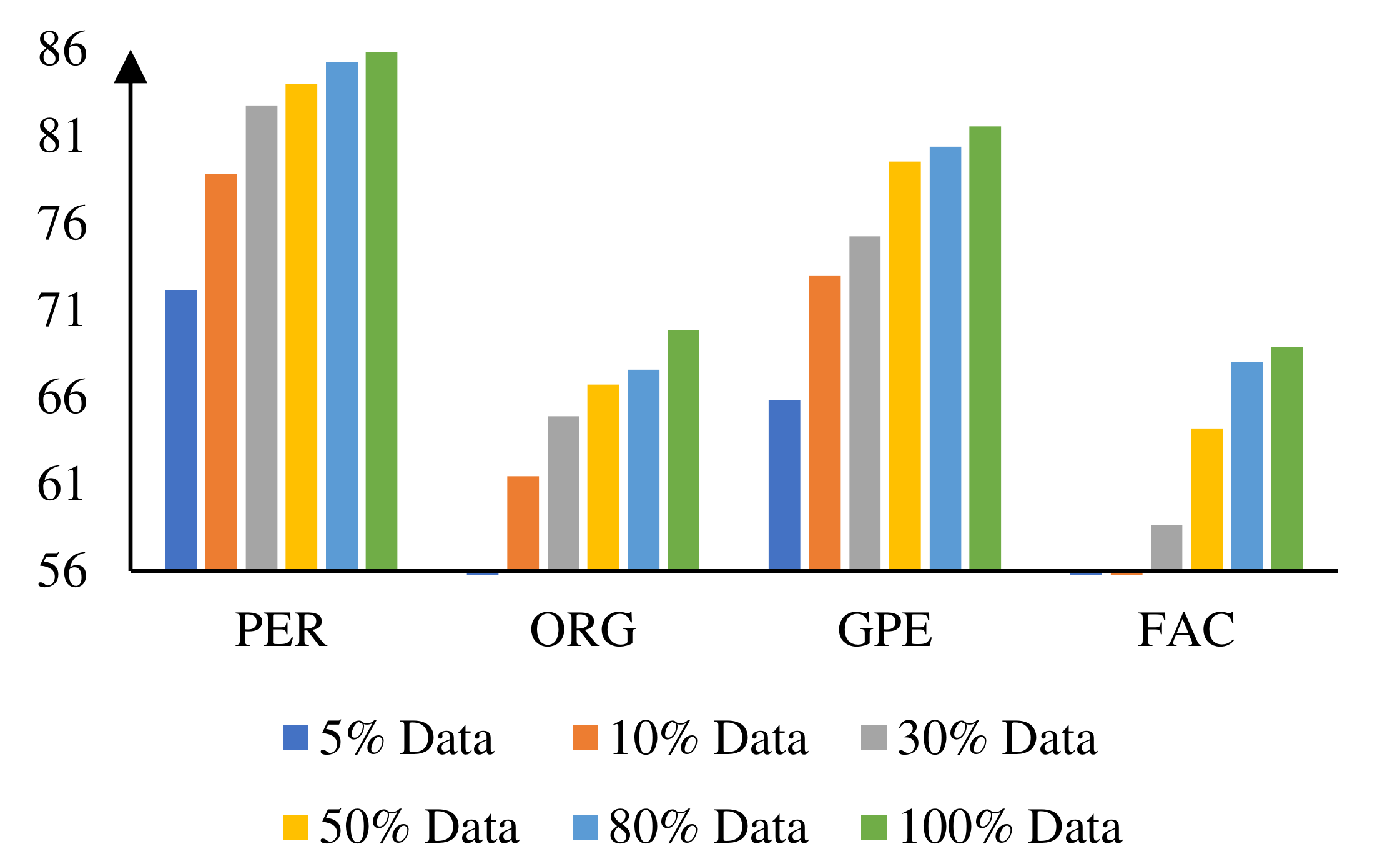}
        \caption{Sentence Reduction}
      \end{subfigure}%
      \begin{subfigure}[c]{0.33\textwidth}
        \centering
        \includegraphics[width=\textwidth]{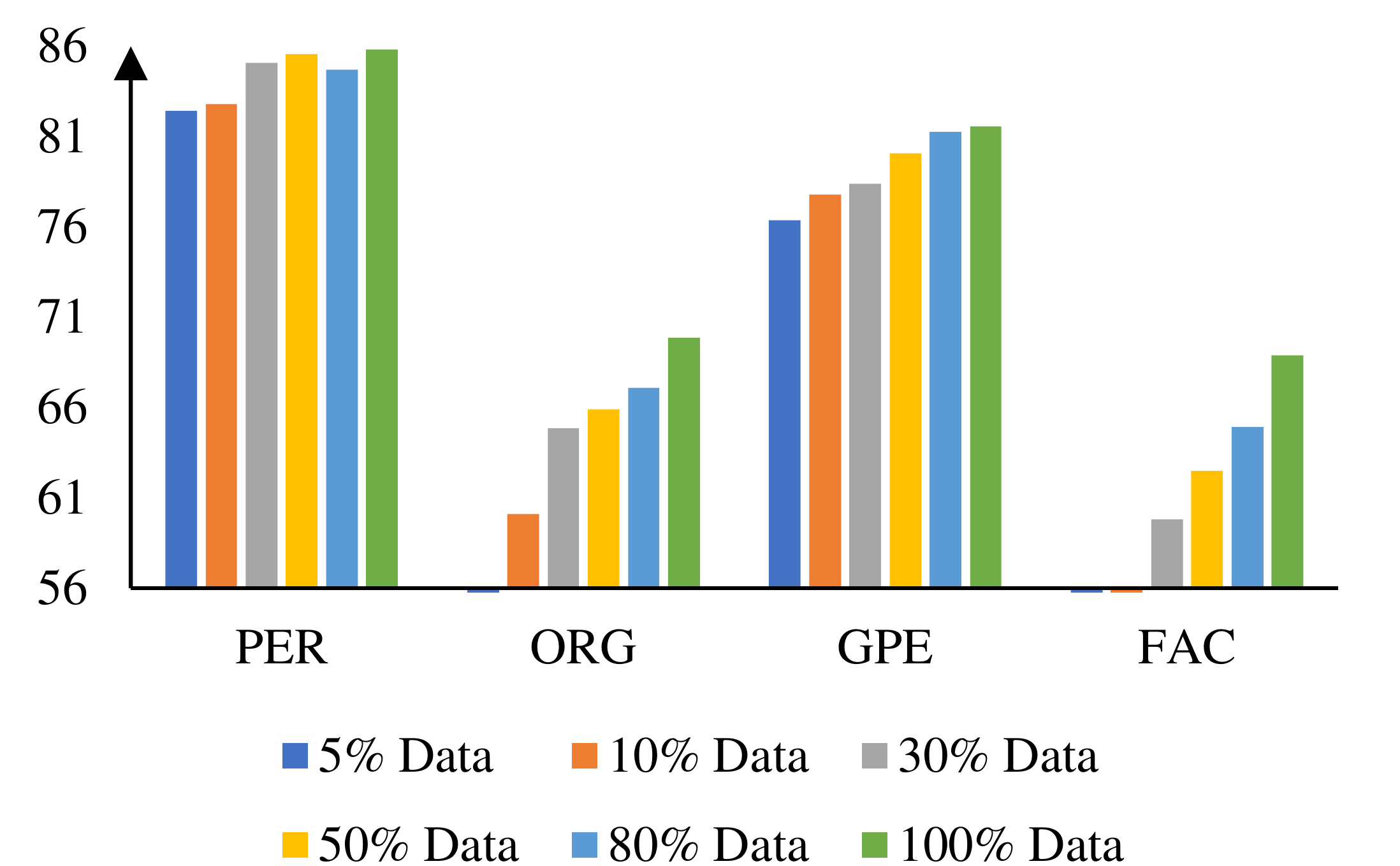}
        \caption{Mention Reduction}
      \end{subfigure}
  }
  
  \caption{Experiments on context reduction, mention reduction and sentence reduction when the kept information ratio varies. We can see that when preserving all name regularity information, sufficient context patterns can be captured once the training sentences reaches a certain amount, and introducing more training data does not significantly improve the performance.} 
  
  \label{fig:context_rep}
  \end{figure*} 
  
Current NER benchmarks commonly provide decent training data for learning context patterns of entities. However, due to the expensive cost, it is usually impractical for open NER to assume enough fully-annotated training data. If we can figure out how many training instances are sufficient for context pattern and name regularity respectively, it will provide valuable insights for constructing open NER datasets and models more effectively and efficiently.

To this end, we conduct \emph{context reduction (CR)} and \emph{mention reduction (MR)} on the vanilla training set using simple data augmentation strategies. The purpose of context reduction is to reduce context diversity in training data but still keeps all name regularity in the vanilla setting. Specifically, CR only keeps a subset of sentences in the vanilla training data, and then duplicates preserved sentences and randomly replace mentions in them with mentions of the same type in the vanilla training data. In this way, all mentions will share identical frequency in the vanilla and CR dataset. On the contrast, MR aims to reduce mention diversity for name regularity, but retains context diversity by keeping all of the original contexts. For this, MR only keeps a part of mentions in the original training set as seeds, and replace other mentions in the training data with a mention randomly sampled from the seeds of the same type. In this way, only part of name knowledge will retain, but all contexts will be preserved. Furthermore, we also compare CR and MR with a naive reduction strategy which simply subsamples sentences in the training data, and we refer it as \emph{sentence reduction}.
  
We varied the ratio of preserved information in each setting ranging from 5\% to 100\% respectively. The overall results are shown in Figure~\ref{fig:context_rep}. We can see that in the sentence reduction setting, the performance steadily improved as the training data grows. This phenomenon is also observed in the mention permutation setting, which indicates that increasing training data will introduce more name regularity knowledge, and thus results in better performance. However, for context permutation setting, there is no significant performance improvement on PER, ORG and GPE when the preserved sentences are more than 30\% of the vanilla data. But for FAC, increasing the preserving data will still improve the performance. This may because the number of FAC instances are significantly smaller than PER, ORG and GPE in the vanilla dataset. From the above experiments, it seems that once it reaches a certain amount, the instances in training data are enough to capture sufficient context patterns. And increasing training instances can mainly provide more name regularity knowledge rather than more context diversity.

The above results provide a valuable insight that the name regularity and the context patterns for NER can be learned separately, rather than jointly. For example, we can learn context patterns using a moderate number of training instances and attempt to incorporate more name regularity knowledge using other resources, e.g., gazetteers.

\section{Experiments on Open NER}

\subsection{Data Preparation}
To further verify the conclusions from our randomization test, we propose to conduct experiments on a real-world open NER dataset, which focuses on real world entity types with weaker name regularity than previously-used benchmarks. Because currently no suitable dataset is available for verifying our conclusions, this paper constructs a new dataset from Wikipedia. Specifically, we consider four entity types in our experiments, including \emph{movie}, \emph{song}, \emph{book} and \emph{tv series}. We extract all sentences in Wikipedia which contain mentions linking to entities of these types as our experiment dataset. From them, we randomly sample 10,000 sentences as the test set and 2,000 sentences as the development set, and part of the remaining data will be used as the training data according to the following different settings. Finally, there are 2875, 2791, 598 and 580 mentions for movie, tv series, song and book in the test set respectively. Note that different from real scenarios, this dataset only keeps sentences containing at least one mention, Due to the partial labeling nature of Wikipedia,  this dataset only keeps sentences containing at least one mention, and the performance on this dataset may over-estimate the precision than in real applications. Although this may different from real scenarios, we believe it can still lead to reasonable conclusions.

\subsection{Generalizing over Unseen Mentions}
The first group of experiments was conducted to verify the influence of name regularity on in-dictionary and out-of-dictionary mentions. To this end, we randomly sampled 5,000 sentences from the dataset as the training set, which is close to the training data size of ACE2005. We use \emph{bert\_cased\_L-24\_H-1024\_A-16} rather than the uncased version of the pretrained model because we find that capitalization may have a significant impact on this dataset.   Furthermore, different from the ACE2005 whose training data covers nearly 58\% test set mentions, the training set of our Wikipedia dataset can only cover 27\% mentions in the test set. This confirms our concern that the mention coverage is much lower in open NER than in regular NER.

\begin{table}[!ht]
  \centering
  \resizebox{0.5\textwidth}{!}{
    \begin{tabular}{l|cc|c||cc|c}
    \hline
    & \multicolumn{6}{|c}{\textbf{Baseline}} \\
    \cline{2-7}
    & \multicolumn{3}{c||}{Precison} & \multicolumn{3}{c}{Recall} \\
    \cline{2-7}
    & InDict & OutDict & \textbf{Diff} &  InDict & OutDict & \textbf{Diff} \\
    \hline
    Movie	&	80.68	&	71.48	&	11\%	&	88.43	&	71.43	&	19\% \\
    Tv Series    &	91.48	&	65.77	&	28\%	&	88.03	&	74.73	&	15\% \\
    Song	&	77.94	&	56.42	&	28\%	&	68.83	&	62.50	&	9\% \\
    Book	&	83.72	&	53.99	&	36\%	&	68.57	&	55.03	&	20\% \\
    \hline
    ALL	    &	87.44	&	66.59	&	\textbf{24\%}	&	86.30	&	71.13	&	\textbf{18\%} \\    
    \hline
    
    \end{tabular}
  }
    \caption{Comparasion between in-dictionary portion and out-of-dictionary portion on Wikipedia dataset. We can see that there is a significant gap between these two portions.}
  \label{tab:wiki_inout_comp}%
\end{table}

Table~\ref{tab:wiki_inout_comp} reports the experiment results on in-dictionary portion and out-of-dictionary portion respectively. We can see that the performance gap between in-dictionary portion and out-of-dictionary portion significant due to the weak name regularity. This confirms our Conclusion 1 that name regularity is vital for NER system to generalize over unseen mentions. Compared with the name permutation setting shown in Table~\ref{tab:full_np}, the InDict-outDict performance gap is not so large. We believe this is because: 1) there still exist some kinds of name regularity for these entity types, e.g., the capitalization of the first letter; 2) Wikipedia documents are much formal than ACE2005 documents, which makes the context patterns much easier to capture. For example, a movie mention in Wikipedia will frequently share the same context of ``in the film Xxx Xxx'', where ``Xxx'' is mention word with the first letter being capitalized. Despite this, the performance gap between InDict and OutDict portions is still significant -- more than 24\% and 18\% on precision and recall respectively, which verifies the necessity of name regularity for NER to achieve good generalization.

\subsection{Influence of Training Data Size}

To investigate the impact of training data size to the model performance, we varied the size of training set from 500 to 10,000, and investigate the performance improvement over the test set. Because the entity types we considered are with weak name regularity, the increment of training instances will mainly increase the context diversity. Therefore, this group of experiments can be used to verify the Conclusion~\ref{con:3} we proposed before.

\begin{figure}[!t]
  \centering
  \includegraphics[width=0.5\textwidth]{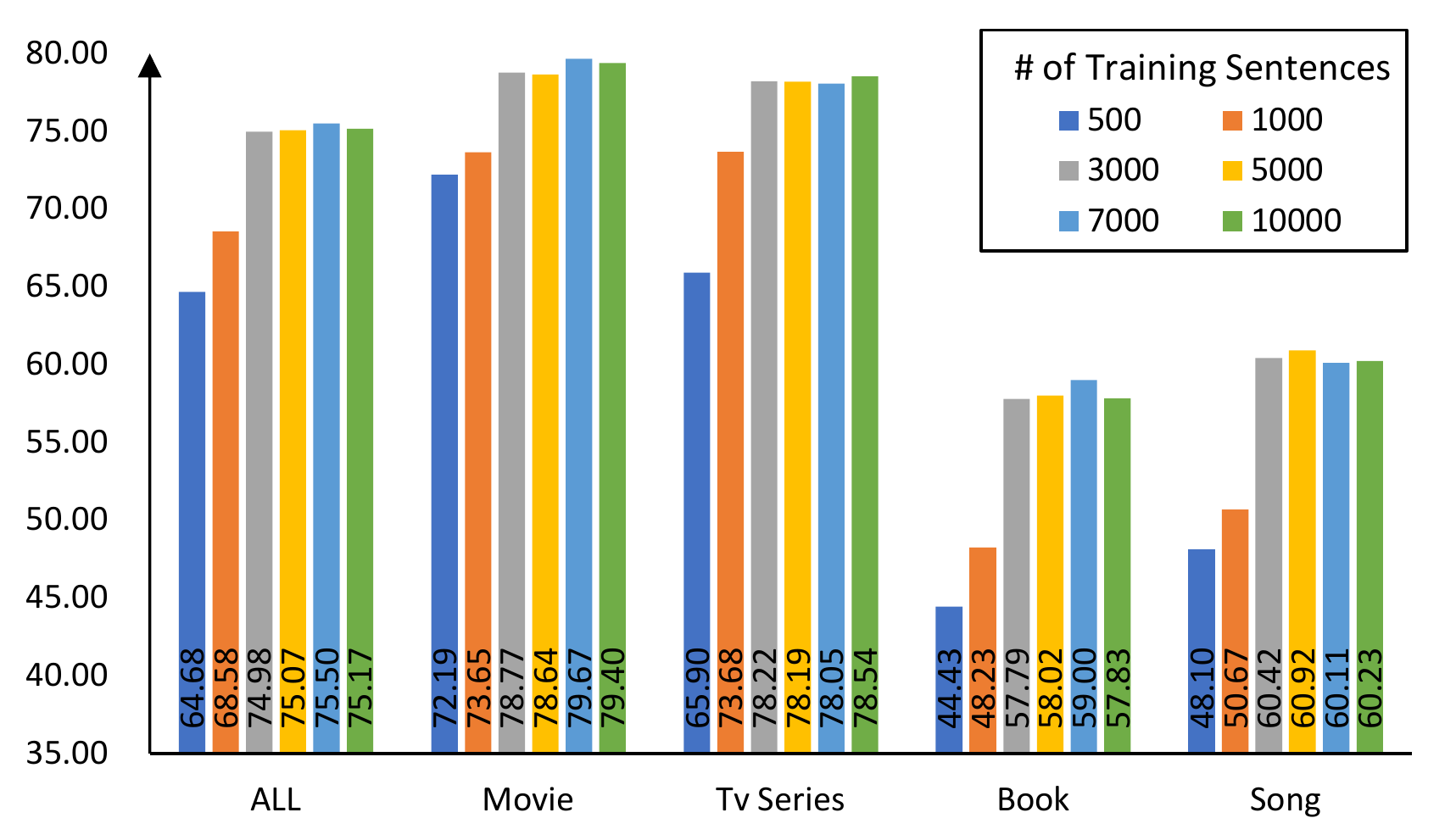}\\
  \caption{F1 scores on Wikipedia dataset when training data size varies. We can see that there are little improvement when training data size exceeds 3000.}\label{fig:wiki_data_size}
\end{figure}

Figure~\ref{fig:wiki_data_size} shows the results. We can see that the performance improvements on all entity types are less significant after training data size exceeding 3000. This phenomenon is to the Figure~\ref{fig:context_rep} (a) results on the context reduction setting on ACE2005. This further verifies our Conclusion~\ref{con:3} and confirms that when sample size reaches a certain level, introducing more training data will not improve the learning of context knowledge.

\section{Related Work}
Named entity recognition has long been studied and has attracted much attention.  Conventional methods~\cite{zhou2002named,chieu2002named,bender2003maximum,settles2004biomedical} commonly rely on handcraft features to build NER models,  which are hard to transfer among different languages, domains and entity types. Recently, deep learning methods, which automatically extract high-level features and perform sequence tagging with neural networks~\citep{santos2015boosting,chiu2016named,lample2016neural,yadav2019survey}, have achieved significant progress especially under strong pretraining and fine-tuning paradigm~\cite{li2019entity,akbik2019pooled,zhai2019improving,li2019unified,xia2019multi,lin2019sequence}. These methods have achieved promising results in almost all popular NER benchmarks considering regular entity types.

Several researches have shift attention to name tagging in open scenarios, where entity types may have weaker name regularity and training data are often insufficient. These papers mainly focus on how to exploit weakly-supervised data~\cite{tackstrom2013token,ni2017weakly,cao2019low,Xue2019NeuralCE}, or devoted to incorporate external resources~\cite{yang2016transfer,peng2016improving,pan2017cross,lin2018multi,xie2018neural,lin2019gazetteer}.

By contrast, to the best of our knowledge, this is the first work which investigates the essential difference between regular and open NER. By conducting both randomization test~\cite{edgington2007randomization} and verification experiments, we analyze the impact of name regularity, mention coverage and context pattern sufficiency and shed light on future open NER studies. 

\section{Conclusion and Future work}
This paper investigates whether current state-of-the-art models on regular NER can still work well on open NER. From the perspective of name regularity, mention coverage and context diversity, we conducted both randomization test and verification experiments to evaluate the generalization ability of models. Our investigation leads to three valuable conclusions, which shows the necessity of decent name regularity to identify unseen mentions, the hazard of high mention coverage to model generalization, and the redundancy of enormous data to capture context patterns.

The above findings shed light on the promising directions for open NER, including 1) exploiting name regularity more efficiently with easily-obtainable resources such as gazetteers; 2) preventing the overfit on popular in-dictionary mentions with constraints or regularizers; and 3) reducing the need of training data by decoupling the acquisition of context knowledge and name knowledge.

\section*{Acknowledgments}
We sincerely thank the reviewers for their insightful comments and valuable suggestions. Moreover, this research work is supported by National Key R\&D Program of China (2020AAA0105200), the National Natural Science Foundation of China under Grants no. U1936207, and in part by the Youth Innovation Promotion Association CAS (2018141).

\bibliography{ACL20}
\bibliographystyle{acl_natbib}

\end{document}